\documentclass[11pt]{article} 

\usepackage{graphicx}
\usepackage{booktabs}
\usepackage{algorithm}
\usepackage{algpseudocode}
\usepackage[T1]{fontenc}
\usepackage{lmodern}
\usepackage[flushleft]{threeparttable}
\usepackage[hidelinks]{hyperref} %
\hypersetup{hidelinks}
\usepackage{multirow}
\usepackage{longtable}
\usepackage[margin=3cm]{geometry} %
\usepackage{amsmath}
\usepackage{authblk}
\usepackage{caption}

\title{Using predefined vector systems as latent space configuration 
for neural network supervised training on data with arbitrarily large number of classes}
\author[1]{Nikita Gabdullin}
\affil[1]{Joint Stock "Research and production company "Kryptonite" \authorcr
E-mail: n.gabdullin@kryptonite.ru}
\date{}
\begin{document}

    \captionsetup[table]{labelformat={default},labelsep=period,name={Table}}

    \maketitle

    \begin{abstract}
        Supervised learning (SL) methods are
        indispensable for neural network (NN) training used to perform
        classification tasks. While resulting in very high accuracy, SL
        training often requires making NN parameter number dependent on the
        number of classes (\emph{n\textsubscript{classes}}), limiting their
        applicability when \emph{n\textsubscript{classes }} is extremely large or
        unknown in advance. In this paper we propose a methodology that allows
        one to train the same NN architecture regardless of
        \emph{n\textsubscript{classes}}. This is achieved by using predefined
        vector systems as the target latent space configuration (LSC) during NN
        training. We discuss the desired properties of target configurations and
        choose randomly shuffled vectors of \emph{A\textsubscript{n}} root
        system for our experiments. These vectors are used to successfully train
        encoders and visual transformers (ViT) on Cinic-10 and ImageNet-1K in low- 
        and high-dimensional cases by matching NN predictions with the predefined v
        ectors. Finally, ViT is trained on a dataset with
        1.28 million classes illustrating the applicability of the method to the
        extremely large \emph{n\textsubscript{classes}} dataset training. In
        addition, potential applications of LSC in lifelong learning and NN
        distillation are discussed illustrating versatility of the proposed
        methodology.
  
    \end{abstract}

    
    \section{Introduction}
\label{introduction}

Modern day technology depends greatly on neural networks (NNs). NNs are
widely applied in many fields including computer vision (CV), autonomous
driving, cybersecurity, manufacturing, healthcare, and others. The growth in the
amounts of data the NNs are expected to process has led to substantial
increase in NN model size and associated computational costs. This can
be illustrated by the rapid increase in model parameter numbers from
multimillion to multibillion in recent years~\cite{4o,Mega}. This is
related to the necessity to analyze high variance data and produce high
quality features that can account for the nuanced differences withing
the data.

The latter capability of NNs is associated with their discriminative
ability which determines how representative NN embeddings, or the
outputs models produce, are. It is partly determined by the quality of
the embedding distribution in NN latent space (LS), where similar input
embeddings must be closer than dissimilar ones~\cite{CenterLoss}. This
requirement has motivated researchers to develop training methods that
ensure good clusterization of similar embeddings and separation of
different clusters, which is commonly achieved by adding specialized
loss functions.

In supervised learning (SL) this often comes at the expense of having
some NN layer's size depend on the number of classes (\emph{n\textsubscript{classes}}). 
This is also the case for training purely with classification losses, i.e. with
cross-entropy (CE) loss. However, many real-life NN applications require
very large \emph{n\textsubscript{classes}} which
also might change during its lifetime making conventional training
methods inefficient. This is also relevant for self-supervised learning
(SSL) where desired cluster numbers can be even higher. This motivates
the search for a training methodology which would not require
associating NN size with the number of classes or clusters while
ensuring that the desired embedding distribution in LS is achieved.

The possibility of obtaining a predefined embedding cluster distribution
for CE-combined training has previously been shown using a methodology
named latent space configuration (LSC)~\cite{lsconf}. There LSC was used
in combination with CE loss to configure embeddings of person
reidentification NN used for classification and similarity ranking~\cite{reid-review}. 
In this paper we formalize the LSC methodology showing its
potential as a stand-alone training method. We discuss the general
approach to the target embedding configuration choice and suggest
possible configurations. We verify that NN training using predefined
embedding distribution is possible without special classification loss
functions, e.g. CE, by conducting experiments on small datasets using
NNs with low-dimensional embeddings. We then extend these ideas to
scenarios with conventional architectures with high-dimensional
embeddings trained on large datasets, e.g. ViT~\cite{VT} trained on
ImageNet-1K dataset~\cite{IN}. We discuss the differences in training
using low/high numbers of LS dimensions (\emph{n\textsubscript{dim}}),
and illustrate the applicability of the proposed method to training on
data with extremely large \emph{n\textsubscript{classes}}. The latter is
possible due to the absence of the direct dependence of NN parameter
number on \emph{n\textsubscript{classes}} so only small batches of
target configuration vectors are used during training.

The rest of the paper is organized as follows: Section~\ref{LSloss} provides an
overview of relevant SL and SSL methods that control embedding
distributions, Section~\ref{methodology} outlines the LSC methodology, Section~\ref{embstudy} studies
the embedding distributions of NNs trained with SL methods, Section~\ref{exp}
provides LSC experimental results, Section~\ref{discussions} discusses different aspects
and application scenarios of LSC, and Section~\ref{conclusions} concludes the paper.

\section{Loss functions acting in LS}
\label{LSloss}

\subsection{Supervised methods}
\label{SL}

The topic of controlling the embedding distribution during NN training
has received a lot of attention in CV research literature of the last
decade. While the major part of this research has been done by the face
recognition community, the training approaches and their underlying
ideas are far more general. It has been shown that to achieve high
discriminative ability, NN embedding clusters should have two important
properties: low intra- and higher inter-class variances~\cite{CenterLoss, SphereFace}. 
This ensures linear separability of the embeddings which
is desired for classification tasks. This means that embeddings of
similar inputs must be close in LS, and groups of embeddings of
different classes must be well-separated. This is achieved by
introducing loss functions that affect NN embedding distribution
directly by acting in LS. These loss functions commonly work as a
supplement to CE loss which is still the most powerful and widely-used
classification loss function~\cite{DL, CVNN}.

The first loss function that allows to satisfy these requirements is
contrastive loss~\cite{Cont}. It uses positive and negative pairs of
inputs during training to maximize the similarity between the former and
minimize it for the latter. This is most often achieved through entropy
calculations~\cite{Unif}. An interesting feature of this loss function is
the possibility to use augmented views of the same image as positive
pairs making it also applicable for SSL. However, contrastive losses are
very sensitive to hyperparameters and they theoretically perform best
having infinite negative samples.

The negative sample choice becomes a significant issue of contrastive
methods in practice. Since it is not feasible to match every positive
pair with all possible negative samples, loss functions of this type
suffer from the ambiguity of the negative sample choice. Moreover, this
choice can significantly affect NN performance, giving rise to various
methods known as negative pair mining~\cite{Sampling, FN}. Nevertheless,
it remains an open problem and one of the main drawbacks of contrastive
methods.

Another popular application of contrastive principle in CV is triplet
loss, which uses triplets of samples (called anchor, positive, and negative)
and minimizes the distance between same class (anchor and positive) pair
embeddings while pushing the others away~\cite{FaceNet}. Triplet loss is
at heart of, for instance, Siamese networks~\cite{Siam}. Despite their
initial success, they also suffer from the negative sample choice
problem due to the large number of possibilities and their influence on
training and the overall NN performance. Training with triplet loss is
also computationally demanding since it requires multiple NN model
copies.

An alternative method is working directly with clusters and their
distribution rather than embedding pairs or triplets. Center loss has
been proposed as a method of determining the optimal cluster
distribution by making cluster centers learnable NN parameters~\cite{CenterLoss}.
It achieves cluster compactness by penalizing LS
Euclidean distance between embeddings and their corresponding cluster
centers. The total training loss is a combination of CE loss and center
loss. This approach has been shown to improve embedding clusterization
and the overall NN performance on face recognition tasks. However,
making cluster centers trainable parameters requires having a NN layer
with size equal to the number of classes or clusters. This limits center
loss application in tasks that require extremely large numbers of
classes.

Another notable class of loss functions are margin losses that are
designed to ensure the desired separation between clusters. These
approaches introduce a specific margin that separates different classes
while grouping same class embedding together. Notable examples of margin
losses include ArcFace~\cite{ArcFace}, CosFace~\cite{CosFace} and,
SphereFace~\cite{SphereFace}, which have shown impressive results on face
recognition tasks. Whereas the idea of cluster separation in LS seems to
be of interest to our study, the exact implementation of this idea used
in margin losses is related to modified log-likelihood distribution
calculation (similar to CE) and not to some constraints applied to
embeddings or LS itself. Therefore, margin losses are not relevant to
the experimental work discussed in this paper.

Finally, prototypical NNs (PNNs) have been proposed as a solution to
several problems in few- and zero-shot learning~\cite{proto, gauss}.
These tasks require classifiers to be able to generalize to new classes
not seen during training by having only a few or no examples of new
class data~\cite{fewshot, zeroshot}. PNNs are relevant to this study
because they associate classification with distance calculation between
target's embedding and precomputed class prototypes. Class prototypes
are defined as mean embeddings of every class calculated on a support
set (a small labeled dataset)~\cite{meanembs}. Class labels for query
images are then calculated as softmax over distances between query's
embedding and all class prototypes, which is very similar to the
approach discussed in Section~\ref{metric}. However, PNN class prototypes are not
predefined which makes them dependent on how representative the data
used for the support set is. PNNs also requires computing softmax over
distances to all class prototypes, which is ineffective in case of very
large \emph{n\textsubscript{classes}}.

\subsection{Self-supervised methods}
\label{SSL}

Good clusterization and separation of embeddings is desirable when
training with SSL methods, too. Having well-defined clusters is
particularly beneficial in SSL since it allows to apply k-means~\cite{kmeans}
or k-NN~\cite{kNN} during training or inference. It has been
shown that preferred clusters can be learned when the predictions from
earlier training epochs are used as pseudo-labels for subsequent epochs~\cite{DeepCl}.
Specifically, k-means is used to obtain a centroid matrix
(a matrix of mean embeddings) and minimize the Euclidean distance
between embeddings and corresponding centroids. While being
methodologically important for its time, the features obtained with this
method are not of desirable quality, and the necessity to alternate
between training and clusterization phases increases the computational
burden of the method.

An important problem in SLL is avoiding trivial solutions (or
representation collapse) when NN outputs the same prediction regardless
of the input. In this paper we are interested in methods that avoid
representation collapse by directly working with embeddings.
Specifically, this problem can be addressed by employing a momentum
encoder with some form of teacher-student training~\cite{st}. Using
momentum encoders was first proposed in~\cite{MOCO} where it was used to
assist with the negative sample mining problem in contrastive learning.

Momentum encoder was then used to bootstrap embeddings in BYOL~\cite{BYOL}. 
This work is a significant milestone in SSL since it has
shown that negative samples are not necessarily needed. It is sufficient
to use a teacher-student training with teacher network weights being
updated as a moving average or student weights. Differently augmented
views of the same image are provided to teacher and student networks
with loss function being the mean average error (MAE) between their
normalized predictions. In other words, student network is trained to
match embeddings predicted by the teacher network. This simultaneously
prevents the representation collapse and allows to obtain useful
features. The use of augmentation also makes this method data-efficient
while simultaneously improving generalization.

An important continuation of these ideas is DINO~\cite{DINO}. It
implements a somewhat simpler version of student-teacher training and
can be used to obtain embeddings that are separable with cosine distance
(see Section~\ref{metric}). In contrast to other mentioned methods, DINO treats
embeddings as pseudo-class probabilities which allows CE to be used for
training. This also allows k-NN to be used for classification or ranking
tasks~\cite{meanembs,kNN}. Cluster centroids are obtained as average
embeddings of all samples representing a certain class, and weighted
voting is used to produce labels for new unseen data. However, this
comes at a cost of increased parameter number as conventional backbone
embedding (e.g., ViT small (ViT-S) 384-dimensional embeddings) are
upscaled with DINOHead layer making the resulting embedding size several
orders of magnitude larger. Specifically, the authors have found
65536-dimensional embeddings to perform best on ImageNet-1k which only
has 1000 classes. Therefore, it is unclear how the parameter number will
grow when the number of classes increases beyond conventional CV
benchmark experiments.

Finally, I-JEPA takes an approach similar to DINO but uses Euclidean
distance as metric~\cite{ijepa}. Parts of the image are randomly masked
during training and NN is trained to predict the embeddings of the
masked regions. It uses context and target encoders similar to
teacher-student approach, including having a momentum encoder in one of
the branches. It should be mentioned that both DINO and I-JEPA use
augmentation applied to image patches, which makes them particularly
useful for ViTs. However, the necessity to have two NN model copies and
complex augmentation and masking complicate training when compared to
other methods.

\subsection{Image augmentation}
\label{imaug}

Input augmentation, and specifically image augmentation in CV, is an
extremely useful technique that allows to increase data variability by
generating new samples from existing ones~\cite{AlexNet}. It is also
widely used in SSL as a mean of obtaining image views which have similar
target embeddings for training. Furthermore, image augmentation is
essential to ensuring high generalization of NNs~\cite{HVT,oh}.

In this paper we show that LSC training on augmented images is possible,
meaning that LSC potentially is capable of high generalization and can
be applied in SSL setting. In our experiments we use Random Augmentation
method proposed in~\cite{aug} which achieves high augmentation
variability by choosing several random augmentations with random
parameters from a list of possible options. The augmentation choice is
made separately for every batch leading to augmentation variations
across different epochs. Specifically, we use aug\_light1 version
following the methodology previously proposed in~\cite{HVT}.

\section{Latent space configuration methodology}
\label{methodology}

\subsection{The main LSC principles}
\label{genform}

Section~\ref{LSloss} shows that both SL and SSL methods widely use the idea of
minimizing the distance between embeddings and corresponding cluster
centroids with respect to some metric. In this Section we formulate LSC
as a NN training methodology which allows one to obtain a predefined
distribution of embedding cluster centers. Cluster center vectors
\emph{C} are obtained from a generating function, where the number of
possible vectors depends on LS dimension \emph{n\textsubscript{dim}}

\begin{equation}
	C_{dim} = f_{gen}(n_{dim}),
	\label{eq:C_dim}
\end{equation}
\unskip

The details regarding generating functions and their specific examples
are discussed in the next Section. Center vectors that are actually used
for training are then chosen from all possibilities to match the desired
number of clusters. Assuming that for a supervised task the numbers of
clusters and classes coincide, we obtain

\begin{equation}
	C_{classes} = f_{choice}(C_{dim}, n_{classes}).
	\label{eq:C_classes}
\end{equation}
\unskip

Finally, LS distance minimization function (LSC loss) used as the target
loss is

\begin{equation}
	L_{LSC} = f_{dist}(C_{classes}, z),
	\label{eq:L_LSC}
\end{equation}
\unskip

where \emph{z} are NN embeddings. It should be noted that metric choice
matters, since the same vector distribution can have different
properties depending on the distance calculation method. However, a
similarity search can be conducted for any reasonable metric as distance
minimization task between embeddings. Optionally, a label function that
allows to produce labels based on embedding proximity to center vectors
can be chosen

\begin{equation}
	y = f_{label}(C_{classes}, z).
	\label{eq:y_label}
\end{equation}
\unskip

The specific functions used in this study are discussed in Section~\ref{metric}.

In this paper we focus on LSC in a supervised setting where labels are
used to choose which input embeddings correspond to which cluster
centers. The crucial difference of the proposed approach with the
similar methods discussed before is that the center embedding vectors
are predefined, and no NN parameters directly depend on
\emph{n\textsubscript{classes}}. The former means that no contrastive
loss or negative sampling is needed for training because center vectors
are chosen so that the separation between vectors is sufficient. The
latter allows one to train the same NN with LSC on datasets with
arbitrary numbers of classes without increasing NN size. This becomes
crucial when training NNs for large \emph{n\textsubscript{classes}}, as
will be further discussed in Section~\ref{nodep}.

\subsection{Center vector generating functions}
\label{centerfunc}

It has been discussed in Section~\ref{LSloss} that low itra- and high inter-class
embedding variances are desired to achieve good NN performance. In this
paper we approach this task by choosing a predefined distribution of
vectors which will allow us to obtain these properties.

In theory, sampling vectors using a uniform distribution of points on an
\emph{n}-dimensional unit sphere should be the best source of center
vectors because such vectors would have maximum possible separation (and
hence, maximum inter-class distances) for a given number of vectors.
Empirical evidence that the uniform distribution of embeddings improves
NN performance can be found in~\cite{Unif}. However, obtaining a general
solution for a uniform point distribution in \emph{n}-dimensional case
is a well-known open problem related to Thomson problem in physics~\cite{Thom}.
In practice, it can be solved numerically using potential
energy considerations~\cite{Energy, MinH}, e.g. a Gaussian potential
formulation~\cite{Unif}.

However, rather than evenly distributing a given number of points on a
hypersphere we ask how many points can be evenly distributed on a
hypersphere in a given dimension \emph{n\textsubscript{dim}}. This is
closely related to kissing spheres or Tammes problem~\cite{SpherePacking, Packing}, 
which also is a complex problem without a general solution.
Hence, we relax the requirement even further and look for known
\emph{n}-dimensional vector systems that have good separation between
vectors. Fortunately, such systems do exist and in this paper we study
the possibility of training NNs to obtain embedding distributions
matching the distribution of vectors in \emph{A\textsubscript{n}} root
system.

\subsubsection{Properties of \emph{A\textsubscript{n}} root system}
\label{An}

A root system is a specific configuration of vectors in Euclidean space
which satisfies certain geometric conditions~\cite{SpherePacking, Anref}.
This is an important mathematical concept which is closely related to
Lie groups and Lie algebras~\cite{Lie}. However, for the purposes of this
paper we are only interested in vector systems and their geometric
properties. Specifically, we study the applicability of
\emph{A\textsubscript{n}} root system where \emph{n} is the space
dimension. \emph{A\textsubscript{n}} is chosen because it produces a
uniform vector distribution with all vectors having 60° angles relative
to their neighbors. \emph{A\textsubscript{n}} properties also do not
depend on \emph{n}, which makes it reliable in high-dimensional
scenarios. For a given \emph{n}, \emph{A\textsubscript{n}} vectors
(called roots) are basis unit vector \emph{e} combinations in
\emph{(n+1)}-dimensions

\begin{equation}
	\alpha_{i} = e_{i} - e_{i + 1}, i = 1,...,n.
	\label{eq:roots}
\end{equation}
\unskip

\begin{figure}[b]
	\centering
	\includegraphics[scale=0.45]{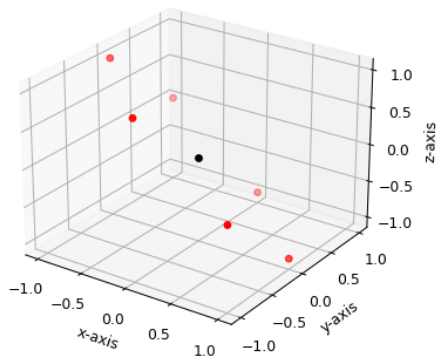} 
	\caption{Visualization of 3-dimensional root vectors of
	\emph{A\textsubscript{2}} located on a 2D plane in 3D space.}
	\label{fig:321}
\end{figure}

Figure~\ref{fig:321} shows the distribution of 3-dimensional root vectors of
\emph{A\textsubscript{2}}. One can see that all vectors lie on a plane
in 3D space. While indeed separated by 60°, such distribution seems
extremely suboptimal since only a 2D plane in a 3D space is occupied.
Furthermore, from an application standpoint the vector space dimension
is NN LS dimension determined by its architecture and cannot be adjusted
freely.

\unskip

\begin{figure}[b]
	\centering
	\includegraphics[scale=0.35]{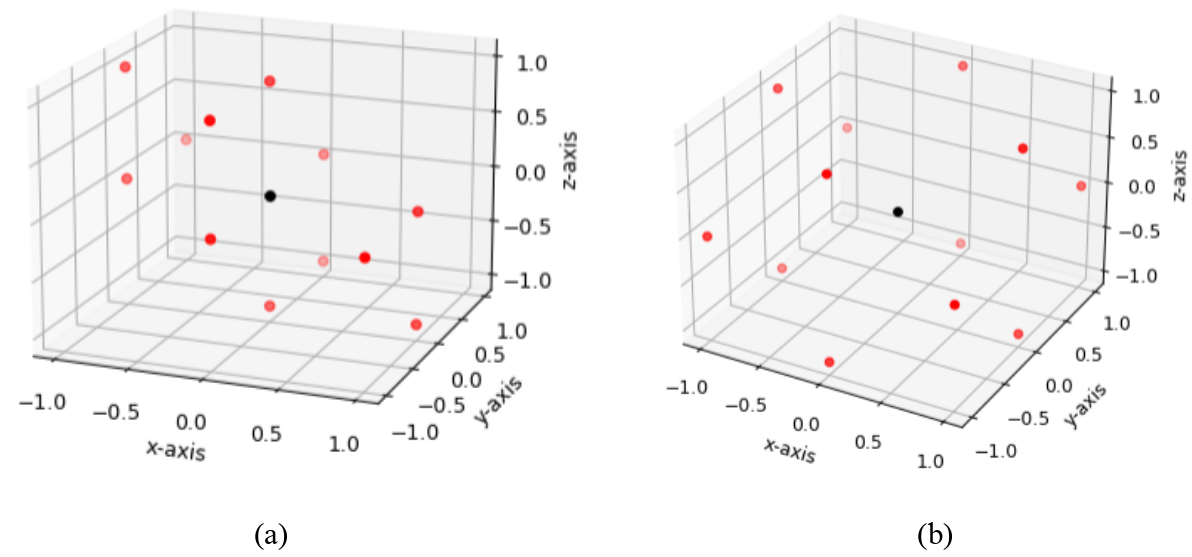} 
	\caption{Root vectors of \emph{A\textsubscript{3}} (a) projected by dropping the
	4\textsuperscript{th} coordinate, and (b) projected with a
	uniformity-preserving operator.}
	\label{fig:322}
\end{figure}

Therefore, we use \emph{n}-dimensional \emph{A\textsubscript{n}} vectors
obtained by projecting the \emph{(n+1)}-dimensional vectors into
\emph{n} dimensions. The simplest way to achieve this is by dropping the
\emph{(n+1)} dimension for all vectors. However,
\emph{A\textsubscript{n}} root vectors are distributed uniformly only in
\emph{(n+1)} dimensions, so this projection operation disturbs the
uniformity of the distribution. It is possible to find a projection
operator that preserves the uniformity using Gram-Schmidt method~\cite{GS}.
However, while the uniformity distortion produced by simple
dimension dropping is prominent in low-dimensional cases as shown in
Figure~\ref{fig:322}, it becomes less and less apparent as \emph{n} grows.
Furthermore, it will be shown in Section~\ref{exp} that distribution uniformity
is not necessarily a desired property. Therefore, we use the
\emph{(n+1)} dimension dropping projection method to obtain
\emph{A\textsubscript{n}} vectors for
\emph{n}=\emph{n\textsubscript{dim}} in our experiments.

\unskip

Table~\ref{tab:32} summarizes the main properties of configurations used in this
study. \emph{A\textsubscript{np}} is obtained from
\emph{A\textsubscript{n}} by using only positive roots, so produced
vectors span only half of the hypersphere and there are no vectors
directly opposite to each other. \emph{A\textsubscript{nr}} is obtained
from \emph{A\textsubscript{n}} by randomly shuffling all root vectors,
resulting in a less ordered vector distribution, especially when the
number of used vectors (which is equal to the number of classes, see
Section~\ref{exp}) is less than the number of \emph{A\textsubscript{n}} roots in
Table~\ref{tab:32}.

It should be noted that additional points can be obtained by
interpolating between \emph{A\textsubscript{n}} roots. Each
interpolation iteration reduces the minimal Euclidean and angular
distances between neighboring vectors by a factor of two. Table~\ref{tab:32}
shows that the number of interpolated points is very large, illustrating
that LS of dimension \emph{n} can accommodate more vectors than just
\emph{A\textsubscript{n}} roots. However, training NN on interpolated
vectors is more difficult and might require reducing learning rate to
achieve convergence, as discussed further in Section~\ref{largenclasses}.

\begin{table}
	\caption{Properties of LS configurations used in this study.} 
	\label{tab:32}
	\centering
	\begin{tabular}{|c|c|c|c|c|}
	  \hline
	  LS & Number of & Number of & Random order & av/min angle \\
	  configuration & root vectors & interpolated vectors & &   \\ \hline
	  \emph{A\textsubscript{n}} & \emph{n(n+1)} & \emph{n(n\textsuperscript{2} - 1)} & no &
	  60/45\\ \hline
	  \emph{A\textsubscript{np}} & \emph{n(n+1)/2} & \emph{n(n\textsuperscript{2} - 1)/2} & no &
	  60/45 \\\hline
	  \emph{A\textsubscript{nr}} & \emph{n(n+1)} & \emph{n(n\textsuperscript{2} - 1)} & yes &
	  60/45 \\\hline
	\end{tabular}
\end{table}

\subsection{Metric choice, loss and label functions}
\label{metric}

As previously mentioned, LSC NN training is performed by minimizing
embedding distances with their corresponding center vectors. When
Euclidean distance is used as metric, loss function (\ref{eq:L_LSC}) becomes the
distance between embeddings and their predefined center vectors~\cite{lsconf}

\begin{equation}
	L_{G} = \sum_{i}^{n_{c}} \sum_{j}^{b_{s}} f_{d}(\sqrt{\sum_{k}^{n_{d}}(z_{jk}(y_{j}=i)-C_{ik})^2},r_{ci}),
	\label{eq:GLoss}
\end{equation}

where \emph{n\textsubscript{c}} is the number of classes,
\emph{b\textsubscript{s}} is batch size,
\emph{n\textsubscript{d}} is the number of LS dimensions, \emph{i} is class
index, \emph{j} is input sample index, \emph{k} is LS dimension index,
\emph{z\textsubscript{j}} is LS position and \emph{y\textsubscript{j}}
is true label of \emph{j\textsuperscript{th}} sample.
\emph{f\textsubscript{d}} is a distance function defined as

\begin{equation}
	f_{d}(x,r_{c}) = exp(ReLU(x - r_{c})) - 1,
	\label{eq:fd}
\end{equation}

and label (\ref{eq:y_label}) is determined by the least distance to one of the
centers

\begin{equation}
	y_{j} = \text{argmin}(\sqrt[]{{(z_{j}\  - \ C)}^{2}}).
	\label{eq:ydist}
\end{equation}

It is well-known that distance functions work worse in high dimensions~\cite{distn}.
However, multiple loss functions discussed in Section~\ref{LSloss}
successfully use distance metric even in high dimensions. The reason for
this apparent contradiction is that high-dimensional NN embeddings are
often located on a lower-dimensional manifold in LS, so the effects
related to high dimensionality become less drastic. However, this is not
the case when embedding vectors are specifically chosen to evenly occupy
all space. It will be shown in Section~\ref{exp} that distance metrics are not
suitable for our purposes outside low-dimensional cases.

To address this issue, cosine distance is used as an alternative metric
that performs well in high-dimensional cases. For a pair of arbitrary
vectors cosine similarity and cosine distance can be written as

\begin{equation}
	\text{sim}_{\cos}\left( x_{1}{,x}_{2} \right) = \ \frac{x_{1} \cdot x_{2}}{\left\| x_{1} \right\|\left\| x_{2} \right\|},
	\label{eq:simcos}
\end{equation}

\begin{equation}
	\text{dist}_{\cos}\left( x_{1}{,x}_{2} \right){= \ 1\  - \ \text{sim}}_{\cos}\left( x_{1}{,x}_{2} \right).
	\label{eq:simdist}
\end{equation}

In this case the loss function (\ref{eq:L_LSC}) is the average cosine distance
between embeddings and corresponding center vectors, which is expressed
through cosine similarity as

\begin{equation}
	L_{\cos} = \frac{1}{b_{s}}\sum_{b_{s}}^{}{\text{dist}_{\cos}\left( z,C_{b} \right)} = 1\  - \ \frac{1}{b_{s}}\sum_{b_{s}}^{}{\text{sim}_{\cos}(z,C_{b})},
	\label{eq:Lcos}
\end{equation}

where \emph{C\textsubscript{b}} are batches of centers with each vector
matching corresponding embedding using true labels, as shown in
Algorithm~\ref{alg:1}. The predicted label (\ref{eq:y_label}) of
\emph{j\textsuperscript{th}} input is obtained as

\begin{equation}
	y_{j} = {\text{argmax}(\text{sim}}_{\cos}(z_{j},C)),
	\label{eq:ycos}
\end{equation}

Algorithm~\ref{alg:1} outlines the general LSC training loop with batched centers
\emph{C\textsubscript{b}}. It shows that only some of the center vectors
have to be sent to GPU after the relevant ones are gathered based on
labels actually present in the batch. Hence, the size of
\emph{C\textsubscript{b}} cannot exceed batch size, which corresponds to
non-repeating labels in the batch. The possibility to use batches of
center vectors makes it unnecessary to store any
\emph{n\textsubscript{classes}}-size objects in memory for training.
This makes GPU load dependent only on NN and batch sizes. This is
different from training with CE loss where classification layer size
depends on \emph{n\textsubscript{classes}} and has to be stored on GPU
at all times along with other NN parameters. The consequences of these
observations are discussed further in Section~\ref{nodep}.

\begin{algorithm}
	\caption{LSC training loop PyTorch pseudo-code with batched center
	vectors \emph{C\textsubscript{b}}. Operations related to optimizer,
	scheduler, and other axillary elements are omitted.}\label{alg:1}
	\begin{algorithmic}[1]
	\State \textbf{Given} NN model with embedding dimension
	\emph{n\textsubscript{dim}}, dataloader that provides pairs of input
	images \emph{x} and corresponding labels, computation device (cuda,
	GPU)
	\State \textbf{initialize} model, dataloader, C =
	f\textsubscript{c}(f\textsubscript{g}(n\textsubscript{dim}),n\textsubscript{classes})
	(combination of (\ref{eq:C_dim}) and (\ref{eq:C_classes}))
	\For{(x, labels) \textbf{in} dataloader}
	  \State z = model(x.to(device))
	  \State C\textsubscript{b} = gather(C, labels).to(device)
	  \State cosine\_sim = simcos(z, C\textsubscript{b}) (using (\ref{eq:simcos}))
	  \State Loss = 1 -- cosine\_sim.mean() (using (\ref{eq:Lcos}))
	  \State Loss.backward()
	\EndFor
	\end{algorithmic}
  \end{algorithm}

\subsection{Using embeddings for similarity search}
\label{embsim}

In this work we primarily use cosine similarity as our metric due to
high dimensionality of studied problems. Incidentally, cosine similarity
is the primary metric for similarity search in vector databases~\cite{db, vdb}, 
industrial surveillance~\cite{surv}, semantic analysis~\cite{cossim}, 
and other important areas. Equation (\ref{eq:simcos}) also shows
that it is extremely efficient computationally since it only requires
taking a dot product of normalized embeddings, which is well-optimized
for modern GPUs. However, knowing the exact distribution of center
vectors when using LSC provides additional advantages since it allows to
apply advanced search algorithms. For instance, space subdivision
algorithms~\cite{vdb, vretriv} significantly reduce the number of
required computations. Whereas \emph{A\textsubscript{n}} specifically
allows to speed up the search even further, this topic will be discussed
in greater details in the future.

\section{Conventional NN embedding study}
\label{embstudy}

\subsection{NN models and datasets}
\label{datasets}

In this paper two main types of experiments are considered: ones that
study LSC training of NNs with different LS dimensions
\emph{n\textsubscript{dim}}, and ones that study LSC training of
conventional models with predefined \emph{n\textsubscript{dim}}. For
former experiments we use the modified UNET~\cite{unet} encoder described
in~\cite{lsconf} with the final linear layers outputs' \emph{a} and
\emph{b} dimensions adjusted to match the desired
\emph{n\textsubscript{dim}}. Hereafter we simply refer to this model as
the encoder. For the latter, a ViT-S with \emph{n\textsubscript{dim}}=
384 is used~\cite{VT}. AdamW~\cite{AW} optimizer with
10\textsuperscript{-4} learning rate and 10\textsuperscript{-5} weight
decay is used for all experiments unless explicitly stated otherwise.
All models are trained using NVIDIA A100 GPU (40GB).

Datasets used in this study are summarized in Table~\ref{tab:41}. Cifar, cinic,
and i1kp are generally used to study LSC in low dimensions, while i1k is
used for main experiments that show the applicability of the proposed
method to the large-scale dataset training. While cinic
originally included cifar images, here they are removed to avoid trivial
results.

\begin{table}
	\caption{Datasets used in this study.} 
	\label{tab:41}
	\centering
	\begin{tabular}{|c|c|c|c|}
	  \hline
	  Dataset & Abbreviation & \emph{n\textsubscript{classes}} & Train/test
	  set size \\ \hline
	  Cifar-10~\cite{CIF} & cifar & 10 & 60k / 10k\\ \hline
	  Cinic-10~\cite{CIN} & cinic & 10 & 90k / 90k\\ \hline
	  ImageNet-1K (full) & i1k & 1000 & 1.28m / -\\ \hline
	  ImageNet-1K (part) & i1kp & 84 & 108k / -\\ \hline
	\end{tabular}
\end{table}

\subsection{Embedding distribution of classifier NNs}
\label{CEembs}

It is well-known that when classifier NNs are trained with CE loss,
their embeddings prior to classification layers can be distinguished
using angular metrics. This happens because angular separation is
inherently consistent with softmax cross entropy~\cite{CosFace}.
Hereafter we refer to the embeddings of classifier NN with removed
classification layer as \emph{CEembs}.

\begin{figure}[b]
	\centering
	\includegraphics[scale=0.45]{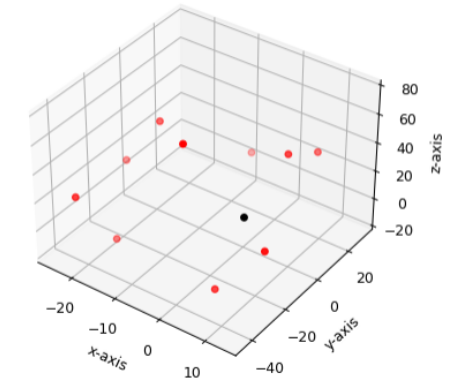} 
	\caption{The distribution of encoder mean embeddings trained on 10
	classes of cinic.}
	\label{fig:421}
\end{figure}

In this section we study the properties of \emph{CEembs} distribution to
see how they relate to the assumptions about good distributions mentioned in
Section~\ref{LSloss} and our target distributions discussed in Section~\ref{An}. Firstly,
an encoder model with \emph{n\textsubscript{dim}}=3 and a single
fully-connected classification layer was trained on cinic. Figure~\ref{fig:421}
shows how its mean class embeddings are distributed in LS. While the
embeddings are separated well, it is hard to make conclusions about the
overall space occupation based on only ten points.

\unskip

\begin{figure}[b]
	\centering
	\includegraphics[scale=0.35]{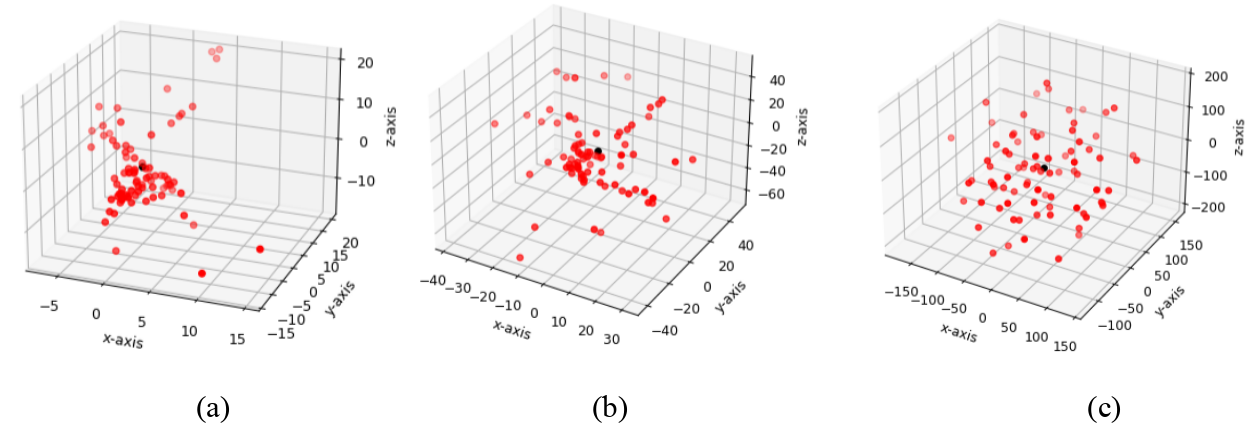} 
	\caption{Embedding distribution (prior to classification layer) of
	the encoder with \emph{n\textsubscript{dim}}=3 trained on i1kp with CE
	loss to (a) 17\% training accuracy on the 1\textsuperscript{st} epoch,
	(b) 35\% training accuracy on 5\textsuperscript{th} epoch, and (c) 96\%
	training accuracy on 60\textsuperscript{th} epoch.}
	\label{fig:422}
\end{figure}
\unskip

\begin{figure}[t]
	\centering
	\includegraphics[scale=0.4]{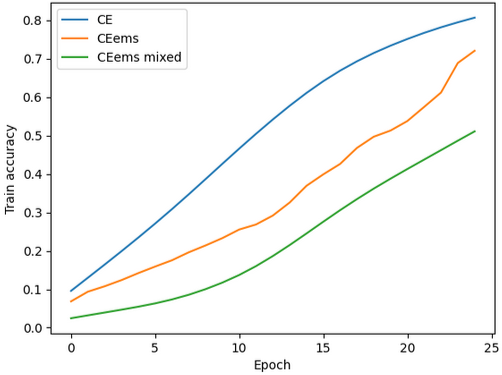} 
	\caption{3-dimensional embedding encoder training speed with CE
	(with a classification layer), and cosine loss using \emph{CEembs} with
	and without mixed labels (without classification layers).}
	\label{fig:441}
\end{figure}
\unskip

Secondly, the same model was trained on i1kp. Figure~\ref{fig:422} shows that
embeddings get distributed more and more evenly as training progresses,
gradually occupying all available space. Indeed, training accuracy is
extremely low when embeddings are crumbled together in Figure~\ref{fig:422} (a),
and it is remarkably high when the embeddings in Figure~\ref{fig:422} (c) are
well-distributed. However, it must be kept in mind that the distribution
in Figure~\ref{fig:422} (c) is non-uniform with some groups of embedding centers
being closer than the others.

However, the comparison of Figures~\ref{fig:421} and~\ref{fig:422} (c) shows that class
mean embeddings get closer as \emph{n\textsubscript{classes}} increases.
This is accompanied by the decreasing average cosine distance between
mean embedding pairs, too. While seemingly trivial, this observation
shows that it is always harder to ensure good embedding separation as
\emph{n\textsubscript{classes}} increases for a fixed
\emph{n\textsubscript{dim}}. This effect is also observed in LSC
training experiments discussed in Section~\ref{exp}.

\subsection{Training by embedding matching}
\label{embmatch}

Remarkably, \emph{CEembs} can be used as a target configuration for LSC
training following the steps discussed in Section~\ref{methodology}, when \emph{CEemb}
vectors are used as target center vectors. Table~\ref{tab:431} shows that
\emph{CEembs}-trained NN has a similar performance to the NN that
sourced the embeddings (experiment 1). Incidentally, this also means
that this training method can be regarded as an approach similar to
distillation~\cite{distil} with average embeddings used as the target
without constantly needing the teacher network, which is discussed further in
Section~\ref{LSCdistil}.

It is also significant that generalization accuracy of the
\emph{CEembs}-trained NN is high, too. This indicates that training
solely with cosine loss does not result in poor NN performance, and
NNs trained to high accuracy with LSC can be expected to perform as well
as their CE-trained counterparts.

\begin{table}
	\caption{ViT-S augmented training on cinic (generalization tested on
	cifar) with classification layer and LSC using \emph{CEembs} as target
	configuration.} 
	\label{tab:431}
	\centering
	\begin{tabular}{|c|c|c|c|c|c|}
	  \hline
	  Experiment & Configuration & Augmentation & \emph{n\textsubscript{classes}} & Loss & Accuracy
	  train/gen, \% \\ \hline
	  1 & - & yes & 10 & CE & 99/75\\ \hline
	  2 & CEembs & yes & 10  & cos & 89/63\\ \hline
	\end{tabular}
\end{table}

\subsection{Mixing labels of target embeddings}
\label{mixedlabels}

Another important question about classifier NN embedding distribution is
whether the specific cluster-label correspondence matters. Indeed,
\emph{CEembs} encode the information not only about the distribution and
relative separation of center vectors, but also about the exact
class-center correspondence. To identify whether this correspondence
matters we train NN to match \emph{CEembs} distribution with target
labels randomly mixed.

Table~\ref{tab:441} shows that NN can be trained to high accuracy with both
original and mixed label \emph{CEembs} configurations. However, Figure~\ref{fig:441}
shows that mixed label training is slower than the original one.
This means that the exact center-label correspondence does matter, which
can be a problem when training NNs on labeled data for which the optimal
label-to-cluster correspondence is unknown.

\begin{table}
	\caption{Training accuracy of encoder with 3-dimensional embeddings
	trained with CE (with classification layer) and two CEembs
	configurations.} 
	\label{tab:441}
	\centering
	\begin{tabular}{|c|c|c|c|c|c|}
	  \hline
	  Experiment & Configuration & \emph{n\textsubscript{dim}} &
	  \emph{n\textsubscript{classes}} & Loss function & Train accuracy,
	  \% \\ \hline
	  1 & - & 3 & 10 & cos & 96\\ \hline
	  2 & CEembs & 3 & 10 & cos & 95\\ \hline
	  3 & CEembs (mixed) & 3 & 10 & cos & 95\\ \hline
	\end{tabular}
\end{table}

\section{Experiments}
\label{exp}

\subsection{LSC in low-dimensional case}
\label{lowdim}

\begin{figure}[b]
	\centering
	\includegraphics[scale=0.3]{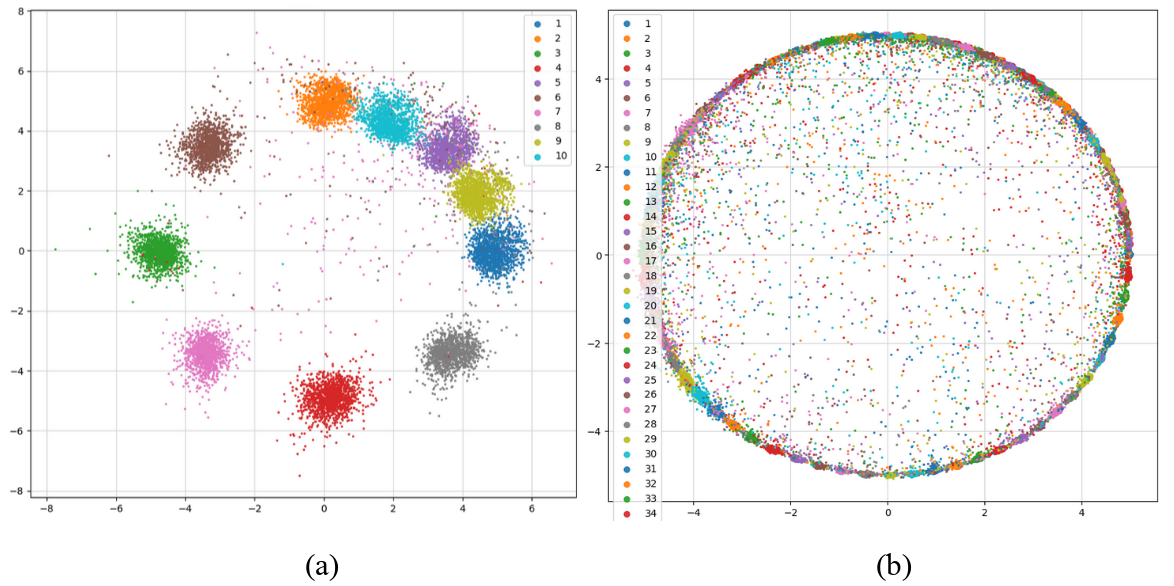} 
	\caption{Training set embedding distribution of encoder model with
	\emph{n\textsubscript{dim}}=2 corresponding to (a) 97\% training
	accuracy on 10 classes of cinic and (b) 89\% training accuracy on 84
	classes of i1kp.}
	\label{fig:511}
\end{figure}
\unskip

We first train encoders in 2D to different numbers of classes
using cinic and i1kp datasets. Only the distance-based loss (\ref{eq:GLoss}) with
\emph{r\textsubscript{c}} = 1 is used for training. For this specific
experiment we do not use \emph{A\textsubscript{2}} root vectors as the
target, but obtain the target center vectors in the following manner.
The first four classes are represented by four vectors that are at 90° angles
relative to each other on a circle with radius \emph{r}=5. Next four
class vectors are obtained by rotating the existing four, which results
in the total of eight vectors at 45° angles relative to the neighbors.
The next eight class vectors are obtained by rotating the existing
eight, and so on. One can see that each rotation operation doubles the
number of vectors and reduces their angular distance by a factor of two.
For each copying-by-rotation operation the cluster size is also reduced
by the factor of two. Whereas this center vector generation method is
useful in 2D, it unfortunately does not scale to high-dimensional cases,
so the \emph{A\textsubscript{n}} roots discussed in Section~\ref{methodology} are used
for all \emph{n\textsubscript{dim}} \textgreater{} 2 experiments
instead.

Figure~\ref{fig:511} shows that the desired distribution is obtained for both 10
and 84 classes. However, whereas cinic training was simple requiring
only 50 epochs, i1kp training took 400 epochs with learning rate
reduction needed after 200\textsuperscript{th} epoch. This happens
because target centers get closer and closer as
\emph{n\textsubscript{classes}} grows in fixed LS dimension, making it
harder for NN to meet the specified requirements. This observation is
consistent with one made for \emph{CEembs} in Section~\ref{embstudy}. Thus, it can be
concluded that higher \emph{n\textsubscript{dim}} is required to
accommodate more classes, and the fewer times we need to interpolate
between existing centers the better. This conclusion holds true to
high-dimensional cases, too.

\begin{figure}[b]
	\centering
	\includegraphics[scale=0.35]{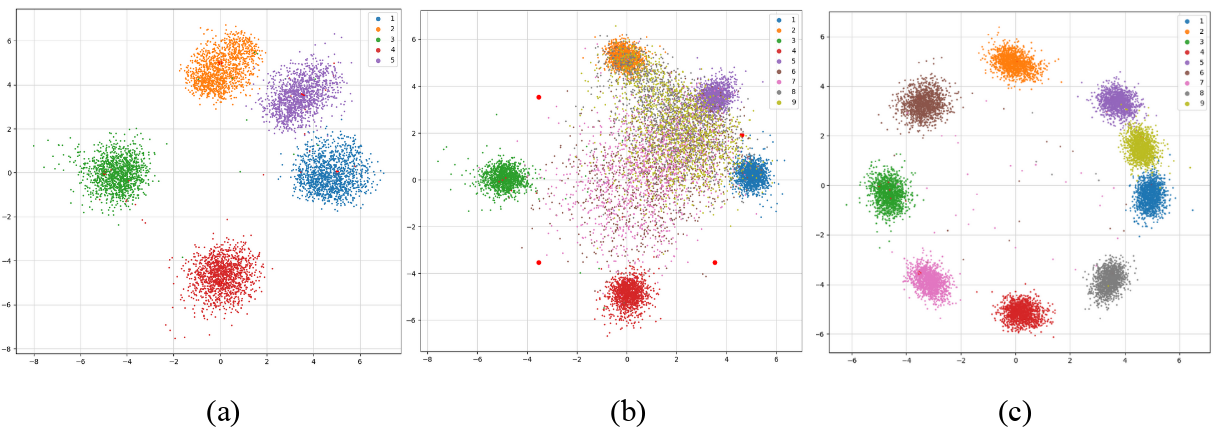} 
	\caption{Cinic training set embedding distribution (a) after
	training using only data of the first 5 classes, (b) at the beginning of
	additional training after adding 6-9 class data, (c) after the training
	on the updated dataset is finished.}
	\label{fig:512}
\end{figure}
\unskip

\subsubsection{Variable \emph{n\textsubscript{classes}} training}
\label{varnclasses}

As it has been shown in Section~\ref{methodology}, \emph{n\textsubscript{classes}} in
LSC appears only in loss function calculation due to its relation to the
number of center vectors. Since NN parameter number is independent of
\emph{n\textsubscript{classes}}, it becomes possible to use the same
architecture for different \emph{n\textsubscript{classes}}. This makes
LSC training useful for tasks that require adding new classes during
operation, e.g. in lifelong or continual learning~\cite{iCarl}.

In this Section we illustrate the possibility of variable
\emph{n\textsubscript{classes}} training while the application scenarios
are discussed in Section~\ref{lifelong}. Figure~\ref{fig:512} illustrates embedding
distribution of encoder model first trained on the first five, and then
on another four classes of cinic. It shows that the desired distribution
is achieved in both cases with no modifications to the model needed when
transitioning from five to nine classes.

The specifics of this experiment are as follows. Firstly, an encoder is
trained on five classes of cinic using loss function (\ref{eq:GLoss}) achieving
the embedding distribution shown in Fig.~\ref{fig:512} (a). Secondly, cinic data
corresponding to classes 6 to 9 is added to the training set. Figure~\ref{fig:512}
(b) shows that as training continues, new unseen class data is
projected somewhere between the existing clusters whereas 1-5 class data
clusters are unaffected. Since in the proposed method LS positions
directly correlate with labels, the predictions for the old data remain
accurate. Such behavior is not guaranteed for conventional classifiers
with fully-connected classification layers which require adding random
weights when changing \emph{n\textsubscript{classes}}.

Finally, the encoder is further trained on the updated dataset. Figure~\ref{fig:512}
(c) shows the final embedding distribution again illustrating that
the desired clusterization has been achieved. This experiment
illustrates that one model can be successfully trained on different and
variable numbers of classes without requiring architecture or parameter
number changes. This is extremely important for tasks where \emph{n\textsubscript{classes}} 
is large, and a few new classes need to be added, and
training for new classes without losing inference performance on the old
classes is needed.

\subsection{LSC in high-dimensional case}
\label{highdim}

\subsubsection{Distance and cosine metrics in high dimensions}
\label{dimmetrics}

Encoder model was trained on i1kp with different
\emph{n\textsubscript{dim}} to study LSC training with different loss
functions in \emph{n}-dimensional case. \emph{A\textsubscript{n}}
configuration was used for all experiments with
\emph{n}=\emph{n\textsubscript{dim}}. Specifically, 9- and
512-dimensional embeddings were chosen. The former allows to accommodate
all 84 classes in \emph{A\textsubscript{9}} configuration without interpolation, and the latter
corresponds to the embedding dimension of CLIP-combined ViT base~\cite{clip}.

Table~\ref{tab:521} shows that whereas both distance and cosine metrics perform
well in 9 dimensions, training with distance loss (\ref{eq:GLoss}) becomes
impossible as \emph{n\textsubscript{dim}} grows. While training with
combined loss improves the performance over distance loss training, pure
cosine training actually allows to obtain the best results. Therefore,
cosine loss is used for all LSC experiments in the following Sections.

\begin{table}
	\caption{Encoder i1kp training accuracy with different
	\emph{n\textsubscript{dim}} and loss functions.} 
	\label{tab:521}
	\centering
	\begin{tabular}{|c|c|c|c|c|}
	  \hline
	  Experiment & \emph{n\textsubscript{dim}} &
	  \emph{n\textsubscript{classes}} & Loss function & Train accuracy,
	  \% \\ \hline
	  1 & 9 & 84 & dist (\ref{eq:GLoss}) & 86.7\\ \hline
	  2 & 9 & 84 & cos (\ref{eq:Lcos}) & 96.1\\ \hline
	  3 & 512 & 84 & dist & 46\\ \hline
	  4 & 512 & 84 & dist + cos & 84\\ \hline
	  5 & 512 & 84 & cos & 98.8\\ \hline
	\end{tabular}
\end{table}

\subsubsection{LSC ViT training}
\label{ViT}

ViT-S with \emph{n\textsubscript{dim}}=384 was trained on i1kp and i1k
to study LSC training of deep models on large datasets. Table~\ref{tab:522}
shows i1kp training results indicating that \emph{A\textsubscript{n}} is
only suitable for training without augmentation, while
\emph{A\textsubscript{np}} works in both cases. This makes the original
\emph{A\textsubscript{n}} system less prospective as NN target
configuration.

\begin{table}
	\caption{Encoder i1kp training accuracy with different
	\emph{n\textsubscript{dim}} and loss functions.} 
	\label{tab:522}
	\centering
	\begin{tabular}{|c|c|c|c|c|c|c|}
	  \hline
	  Exp. & Configuration & Augmentation &
	  \emph{n\textsubscript{classes}} & Loss function & Train
	  accuracy, \% \\ \hline
	  1 & A\textsubscript{n} & - & 84 & cos & 98\\ \hline
	  2 & A\textsubscript{n}  & yes & 84 & cos & -\\ \hline
	  3 & A\textsubscript{np}  & - & 84 & cos & 98\\ \hline
	  4 & A\textsubscript{np} & yes & 84 & cos & 94\\ \hline
	\end{tabular}
\end{table}

However, Table~\ref{tab:523} shows that both \emph{A\textsubscript{n}} and
\emph{A\textsubscript{np}} training attempts fail on i1k even without
augmentation. This raises a question of whether these configurations are
not suitable for large \emph{n\textsubscript{classes}} training. To
answer this question, \emph{CEembs} were extracted from a
classifier-trained ViT (experiment 3 in Table~\ref{tab:523}). Then ViT model
without the classification layer was successfully trained from scratch
using \emph{CEembs} as target configuration (experiment 4 in Table~\ref{tab:523}). 
This indicates that the target configuration but not the number
of classes is the issue.

Inspired by the observation in Section~\ref{embstudy} that \emph{CEemb} distribution
is non-uniform, \emph{A\textsubscript{nr}} with randomly chosen root
vectors was used as the target configuration. This resulted in
successful training both with and without augmentation. This further
verified that non-uniform embedding distributions are preferred by NNs.
While contradicting observations done in~\cite{Unif}, this behavior can
be explained by the fact that some classes are inherently more similar
to each other than others, making a uniform distribution training a more
complex task since is does not account for this effect.

\begin{table}
	\caption{ViT-S i1k LSC training experiments with different target
	configurations.} 
	\label{tab:523}
	\centering
	\begin{tabular}{|c|c|c|c|c|c|c|}
	  \hline
	  Exp. & Configuration & Dataset & Aug. &
	  \emph{n\textsubscript{classes}} & Loss & Train
	  accuracy, \% \\ \hline
	  1 & A\textsubscript{n} & i1k & - & 1000 & cos & -\\ \hline
	  2 & A\textsubscript{np} & i1k & - & 1000 & cos & -\\ \hline
	  3 & - & i1k & yes & 1000 & CE & 89\\ \hline
	  4 & CEembs & i1k & - & 1000 & cos & 89\\ \hline
	  5 & A\textsubscript{nr} & i1k & - & 1000 & cos & 87.9\\ \hline
	  6 & A\textsubscript{nr} & i1k & yes & 1000 & cos & 84.6\\ \hline
	\end{tabular}
\end{table}

\begin{figure}[b]
	\centering
	\includegraphics[scale=0.45]{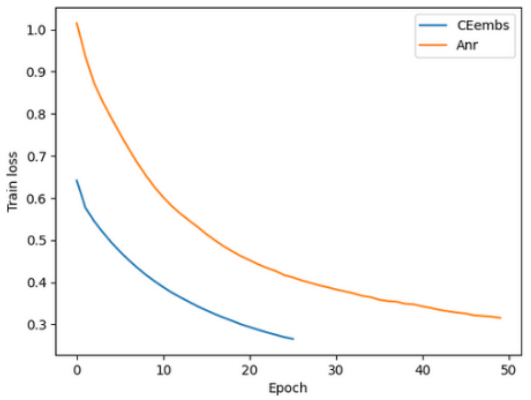} 
	\caption{Training loss curves of ViT-S trained with \emph{CEembs}
	and \emph{A\textsubscript{nr}} target configurations.}
	\label{fig:521}
\end{figure}

Finally, Figure~\ref{fig:521} shows that \emph{A\textsubscript{nr}} training is
slower than \emph{CEemb} training. It has been previously shown in
Section~\ref{mixedlabels} that \emph{CEembs} include information not only about the
distribution, but also the correct label-center correspondence. This
indirectly indicates that \emph{A\textsubscript{nr}} label-center
correspondence might also be suboptimal. While it is theoretically
possible to slightly optimize the target configuration by specifying
which classes should be closer to one another, this is not feasible when
\emph{n\textsubscript{classes}} is extremely large. Addressing this
problem remains an open question in LSC and a possible solution using
SSL is discussed in Section~\ref{discussions}.

\unskip

\subsection{Training on extremely large n\textsubscript{classes} datasets}
\label{largenclasses}

It has previously been mentioned that LSC theoretically allows one to
train NNs on datasets with extremely large
\emph{n\textsubscript{classes}}. However, training on conventional CV
benchmark datasets is computationally demanding while
\emph{n\textsubscript{classes}} they provide, e.g. 21k classes for
ImageNet-21k~\cite{IN} or 83k classes for ms-celem-1m~\cite{MSC}, is
actually not that high. Therefore, training on these datasets would not
prove the applicability to arbitrarily large
\emph{n\textsubscript{classes}}. To address this issue, we train NNs on
an artificial dataset based on i1k which is obtained by assigning a
unique label to each i1k image, resulting in 1.28m unique labels.

Table~\ref{tab:531} shows successful training results for ViT-S and enc. 
In both cases LS dimension size is 384. Base settings
discussed in Section~\ref{datasets} were used for all experiments up to 147k
classes, which corresponds to the number of root vectors of \emph{A\textsubscript{384}}.
Interpolation discussed in Section~\ref{An} was performed to obtain
additional vectors for 300k-1281k experiments. Training on interpolated
vectors also required reducing learning rate to 10\textsuperscript{-5}. 
For experiments with
\emph{n\textsubscript{classes}}\textgreater300k in Table~\ref{tab:531} input
images were resized to 32x32 (which corresponds to input image size of
cifar and cinic) to speed up the computations. However, the possibility
of training ViT-S on original 224x224 i1k images for
\emph{n\textsubscript{classes}}\textgreater300k was verified, too.

\begin{table}
	\caption{Training results of ViT-S and encoder on i1k with artificially
	increased \emph{n\textsubscript{classes}}.} 
	\label{tab:531}
	\centering
	\begin{tabular}{|c|c|c|c|c|c|c|c|}
	  \hline
	  Exp. & Model & Config. & Interpolations &
	  \emph{n\textsubscript{classes}} & \emph{n\textsubscript{dim}} & Loss &
	  Training accuracy, \% \\ \hline
	  1 & ViT-S & A\textsubscript{nr} & - & 10k & 384 & cos & 99 \\ \hline
	  2 & ViT-S & A\textsubscript{nr} & - & 50k & 384 & cos & 98.1 \\ \hline
	  3 & ViT-S & A\textsubscript{nr} & - & 100k & 384 & cos & 96.2 \\ \hline
	  4 & ViT-S & A\textsubscript{nr} & - & 147k & 384 & cos & 92.8\\ \hline
	  5 & enc. & A\textsubscript{nr} & 1 & 300k & 384 & cos & 91.4\\ \hline
	  6 & enc. & A\textsubscript{nr} & 1 & 600k & 384 & cos & 89.2 \\ \hline
	  7 & enc. & A\textsubscript{nr} & 1 & 1281k & 384 & cos & 87.1\\ \hline
	\end{tabular}
\end{table}

\subsection{Optimizing latent space dimension depending on n\textsubscript{classes}}
\label{min_n}

\begin{figure}[b]
	\centering
	\includegraphics[scale=0.45]{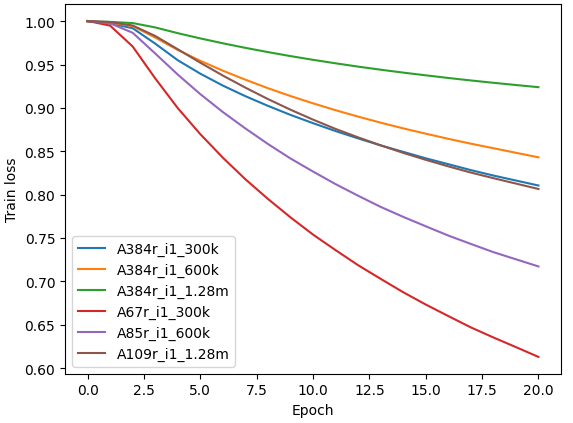} 
	\caption{Training loss curves of encoder trained with with predefined
		\emph{n\textsubscript{dim}}=384 and minimum \emph{n\textsubscript{dim}} that
		corresponds to the desired \emph{n\textsubscript{classes}}.}
	\label{fig:541}
\end{figure}
\unskip

Previous sections explored the possibility of training conventional ViTs
with predefined \emph{n\textsubscript{dim}} on large datasets with
various \emph{n\textsubscript{classes}}. In this case
\emph{n\textsubscript{dim}} determined the number of classes that could
be allocated in LS with and without interpolation according to equations
shown in Table~\ref{tab:32}. However, one could also use the same equations to
find an optimal \emph{n\textsubscript{dim}} for predefined
\emph{n\textsubscript{classes}}.

In this case, for the experiments 5 and 6 in Table~\ref{tab:531}, assuming one
interpolation, one would obtain \emph{A\textsubscript{67r},}
\emph{A\textsubscript{85r},} and \emph{A\textsubscript{109r }}for 300k,
600k, and 1.28m cases, respectively. Figure~\ref{fig:541} shows that training encoders
for minimum \emph{n\textsubscript{dim}} is considerably faster than
training using the original \emph{n\textsubscript{dim}}=384. This means
that using an excessively large \emph{n\textsubscript{dim}} can hinder
NN training when the desired number of classes or clusters is known. The
effects shown in Figure~\ref{fig:541} can likely be explained by easier training
when having low-dimensional embeddings with the same inter-class
distances (which is guaranteed by the same LS configuration). The effects of 
\emph{n\textsubscript{dim}} optimization for other NN architectures will be studied in the future.

\section{Discussions}
\label{discussions}

\subsection{Advantages of having no NN weight dependence on n\textsubscript{classes}}
\label{nodep}

It has previously been discussed in Section~\ref{methodology} that LSC training does not
require linking NN parameters with the number of classes. Conversely,
training a classifier NN requires increasing the classification layer
size proportionally to the number of classes. Table~\ref{tab:611} shows how
classifier NN parameter number increases with
\emph{n\textsubscript{classes}} even for a single fully-connected
classification layer. For instance, it shows that even for
10\textsuperscript{5} classes the classification layer size exceeds the
size of the backbone model. On the contrary, the parameter number stays
constant for LSC training. This makes LSC a possible solution when using
classifiers becomes not feasible or even impossible due to the parameter
number growth.

\begin{table}
	\caption{Parameter numbers, embedding and output sizes of ViT-S with
	and without a fully-connected (\emph{fc}) classification layer for LSC and CE
	training.} 
	\label{tab:611}
	\centering
	\begin{tabular}{|c|c|c|c|c|c|c|c|}
	  \hline
	  Exp. & Method & Model & Loss & \emph{n\textsubscript{classes}} &
	  \emph{n\textsubscript{param}} & Emb size & Out size \\ \hline
	  1 & LSC & ViT-S & cos & 10 & 22m & {[}\emph{b\textsubscript{s}}, 384{]}
	  & {[}\emph{b\textsubscript{s}}, 384{]}\\ \hline
	  2 & LSC & ViT-S & cos & 1000 & 22m & {[}\emph{b\textsubscript{s}},
	  384{]} & {[}\emph{b\textsubscript{s}}, 384{]}\\ \hline
	  3 & LSC & ViT-S & cos & 100k & 22m & {[}\emph{b\textsubscript{s}},
	  384{]} & {[}\emph{b\textsubscript{s}}, 384{]}\\ \hline
	  4 & LSC & ViT-S & cos & 1m & 22m & {[}\emph{b\textsubscript{s}}, 384{]}
	  & {[}\emph{b\textsubscript{s}}, 384{]}\\ \hline
	  5 & Classification & ViT-S + fc & CE & 10 & 22m+384·10 &
	  {[}\emph{b\textsubscript{s}}, 384{]} & {[}\emph{b\textsubscript{s}},
	  10{]}\\ \hline
	  6 & Classification & ViT-S + fc & CE & 1000 &
	  22m+384·10\textsuperscript{3} & {[}\emph{b\textsubscript{s}}, 384{]} &
	  {[}\emph{b\textsubscript{s}}, 10\textsuperscript{3}{]}\\ \hline
	  7 & Classification & ViT-S + fc & CE & 100k &
	  22m+384·10\textsuperscript{5} & {[}\emph{b\textsubscript{s}}, 384{]} &
	  {[}\emph{b\textsubscript{s}}, 10\textsuperscript{5}{]}\\ \hline
	  8 & Classification & ViT-S + fc & CE & 1m & 22m
	  +384·10\textsuperscript{6} & {[}\emph{b\textsubscript{s}}, 384{]} &
	  {[}\emph{b\textsubscript{s}}, 10\textsuperscript{6}{]}\\ \hline
	\end{tabular}
\end{table}

Table~\ref{tab:612} further illustrates the last point showing that ViT base (ViT-B) on
a 40GB NVIDIA A100 GPU cannot be trained for 10 million classes. This is
a consequence of the growth in the memory required to accommodate both
the model and batch data objects. On the contrary, LSC allows to analyze
such cases ensuring that the maximum batch size can be used regardless
of \emph{n\textsubscript{classes}}.

\begin{table}
	\caption{The model and maximum possible batch sizes for 40GB NVIDIA
	A100 GPU training of ViT-B with LSC and conventional classification
	depending on the number of classes.} 
	\label{tab:612}
	\centering
	\begin{tabular}{|c|c|c|c|c|}
	  \hline
	  Experiment & Method & \emph{n\textsubscript{classes}} & Model size, Mb &
	  Max batch size \\ \hline
	  1 & LSC & any & 943 & 386\\ \hline
	  2 & Classification & 1k & 943 & 386\\ \hline
	  3 & Classification & 10k & 973 & 386\\ \hline
	  4 & Classification & 100k & 1237 & 374\\ \hline
	  5 & Classification & 1m & 3873 & 154\\ \hline
	  6 & Classification & 10m & 30281 & -\\ \hline
	\end{tabular}
\end{table}

\subsection{LSC application in lifelong learning}
\label{lifelong}

\begin{figure}[b]
	\centering
	\includegraphics[scale=0.42]{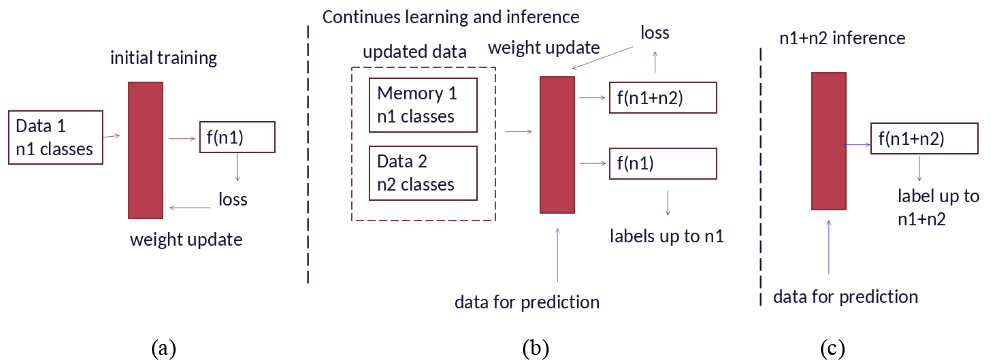} 
	\caption{LSC application scenario in continual learning, (a) model training on the initial dataset with \emph{n\textsubscript{1}} classes, 
	(b) training continuation on additional data \emph{n\textsubscript{2}} with \emph{n\textsubscript{1}}+\emph{n\textsubscript{2}} 
	classes used for training and \emph{n\textsubscript{1}} classes used for inference 
	without losing performance on \emph{n\textsubscript{1}}, and (c) inference on \emph{n\textsubscript{1}}+\emph{n\textsubscript{2}} after 
	training phase in (b) is completed.}
	\label{fig:621}
\end{figure}
\unskip

The ability to learn from new data, called continual or lifelong
learning, is essential in many NN applications~\cite{CLr1}. In CV context
this is related to adding new classes, like new objects in image
classification or new persons in face recognition. The main problem in
this area is catastrophic forgetting, which is related to the model's
performance decrease on old data when training on the new one.
Furthermore, some applications require models to continuously learn from
the environment making the distinction between training and test phases vague~\cite{CLr2}.

Some researchers designed NN architectures capable of dynamically
expanding or allocating weight groups to certain class data to avoid the
forgetting phenomenon~\cite{Prog,DER}. This obviously requires
increasing model size as the dataset size increases, which is a
significant disadvantage. Another approach is using representative
subsets (so-called ``episodic memory'') of old data classes mixed with
new data during continues training~\cite{EWC,iCarl}. This method does
not require modifying the model and relies more on model's
generalization capabilities. In this Section we show that the
independence of model parameters on \emph{n\textsubscript{classes}}
provides additional advantages by guaranteeing correct performance on
old data assuming that episodic memory training allows to alleviate the
forgetfulness effects.

Figure~\ref{fig:621} outlines an application scenario where NN model is first
trained on a dataset with \emph{n\textsubscript{classes}}=\emph{n\textsubscript{1}} 
and then additionally trained on new data with
\emph{n\textsubscript{classes}}=\emph{n\textsubscript{2}}, the scenario
previously discussed in Section~\ref{varnclasses}. Figure~\ref{fig:621} (a) and (c) are the
same for LSC and conventional methods. However, this is not the case for
Figure~\ref{fig:621} (b). When new data is added for conventional methods, there
might be a decrease in performance on old data while the model is
adapting to the sudden parameter change. This makes inference results
for old classes temporary unreliable.

However, the LSC independence of parameters on
\emph{n\textsubscript{classes}} guarantees that no sudden parameter
changes occur since no new parameters are added into the model.
Furthermore, it has been shown in Section~\ref{varnclasses} that old clusters are
not affected when new data is added, meaning that classification metrics
that rely on known cluster positions are unaffected. This guarantees
correct inference results for old data even when model weights are
updated when learning new class clusters.

\subsection{Center vector configuration choice and training limitations}
\label{centerchoice}

In this paper we started our discussion about what constitutes a good
embedding distribution with an overview of different criteria found in
existing research in Section~\ref{LSloss}. These mainly focused on ensuring that
the same-class features are clustered together and different clusters
are well-separated. This then led to the first major assumption that a
uniform distribution of clusters is preferred, with existing research
showing the validity of this assumption~\cite{Unif}. This inspired us to
propose \emph{A\textsubscript{n}} root system as the target distribution
in Section~\ref{An} since it possesses the desired properties in any LS
dimension.

However, we then studied the embedding distributions of conventional
classifiers and discovered that while our assumptions were overall
correct, the distributions one obtains are actually non-uniform. This
observation inspired us to propose random combinations of
\emph{A\textsubscript{n}} vectors (\emph{A\textsubscript{nr}} in
Sections~\ref{methodology} and~\ref{exp}) which worked best when training deep NNs on large
datasets. This distribution currently has performed the best in our
experiments, even though Section~\ref{exp} has shown that training with it is
harder than with NN-preferred \emph{CEembs} indicating that further
improvements are possible.

The success of interpolated \emph{A\textsubscript{nr}} training in
Section~\ref{largenclasses} shows that training using target distributions with vectors
that are closer than \emph{A\textsubscript{n}} root vectors is also
possible. Incidentally, vector systems similar to
\emph{A\textsubscript{n}} which have more base vectors do exist.
Future work will focus on studying such vector systems as potential
candidates for speeding-up LSC training by increasing the efficiency of
LS occupation and better approximating the preferred NN distribution.

However, it is also obvious that choosing an extremely large number of
vectors will result in cluster proximity, making such vector systems
unusable for NN training in practice. This can be illustrated by the
fact that the cosine distance between 30° separated vectors is only
0.134. Training for such fine cluster separation requires NN
architectures with high discriminative ability and low learning rate
optimization, as has been shown in Section~\ref{largenclasses}. Hence, we choose
vector systems which have a large number of well-separated vectors
while avoiding allocating them unreasonably close with respect to our
chosen metric.

\subsection{Controlling intra-class distribution}
\label{intradistrib}

In this paper we have primarily discussed the mutual distribution of
different class clusters and the methods of grouping input data around
the predefined cluster centers. However, there is also a question of the
data distribution within clusters. Some researchers proposed using
Gaussian spheres, or \emph{n}-dimensional Gaussian distribution
approximations, to model target distributions within clusters~\cite{proto}.
This, for instance, can be achieved by matching the mean and standard
deviation (std) of current distribution with a target one. However, the
most common way is using Kullback-Leibler divergence (KLD) to estimate
the difference between two distributions~\cite{vae,distil}.

Approaches that define the target intra-class distribution can readily
be combined with LSC. However, after training NNs with KLD or Gaussian
mean/std matching we have not observed any tangible improvements in the
overall model performance. Figures~\ref{fig:511} and~\ref{fig:512} have previously
showed good clusterization solely with Eucledian distance loss training.
While not shown in this study, similar observations have been made for
angular losses, too~\cite{CenterLoss,lsconf}. Therefore, we conclude
that for the purposes of this study training with distance or cosine
losses is sufficient to achieve good intra-class distributions.

\subsection{Potential application of LSC in NN distillation}
\label{LSCdistil}

\begin{figure}[b]
	\centering
	\includegraphics[scale=0.5]{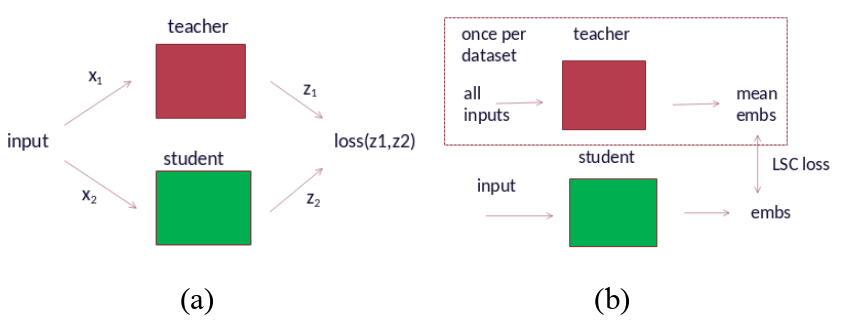} 
	\caption{A comparison of (a) traditional NN distillation with (b)
	LSC distillation approach with a smaller model trained to match the
	embedding distribution of a larger model.}
	\label{fig:651}
\end{figure}
\unskip

Section~\ref{embstudy} has shown that \emph{CEembs} can be used to train one NN to
reproduce the performance of the source NN with the same architecture.
However, embedding matching can be used to train a smaller NN (student)
to operate similar to a larger one (teacher), too, in a manner similar
to NN distillation~\cite{distil}. Figure~\ref{fig:651} illustrates that while
inference of both models is needed for distillation loss calculation,
LSC distillation instead requires precomputing teacher's mean embeddings
on target dataset. Since this operation should be only performed once,
the student training loop can be optimized because the teacher model
does not have to be constantly stored in memory. Furthermore, mean
embeddings can also be efficiently batched as shown in Algorithm~\ref{alg:1}. This
potentially makes LSC distillation faster and more computationally
efficient.

The feasibility of the proposed methods was verified by training an
encoder model using \emph{CEembs} obtained from ViT-S in experiment 3 in Table~\ref{tab:523}.
Similar to the results in Section~\ref{exp}, the performance of the model
trained on \emph{CEembs} (student) was similar to the performance of the source
model (teacher), indicating successful distillation from 22m to 9m
parameters. However, it should be noted that LSC distillation allows
only matching embeddings before the classification layer. Hence, when
one requires logits or hard labels as model output, they would need to
train classification layers separately. Furthermore, methods that work
with precomputed teacher features do exist and a more detailed
comparison with them is required. Conventional distillation also allows
using loss functions which might train faster than the embedding
matching losses proposed in this paper, so the final training speed
ratio depends on multiple effects. LSC distillation will be studied in
greater detail in the future.

\subsection{Obtaining the initial distribution with SSL}
\label{SSLLSC}

It has been discussed in Section~\ref{embstudy} that preferred distributions do exist
for specific combinations of NNs and datasets. It has also been
emphasized that preferred distributions are generally non-uniform. This
makes sense since some classes are inherently more similar than the
others, and NNs would account for such effects differently depending on
architectures. Hence, training to obtain a predefined universal
embedding distribution becomes more difficult depending on how the
target distribution differs from the preferred one. The preferred
cluster-label correspondences are generally unknown until the NN is trained.

\begin{figure}[b]
	\centering
	\includegraphics[scale=0.37]{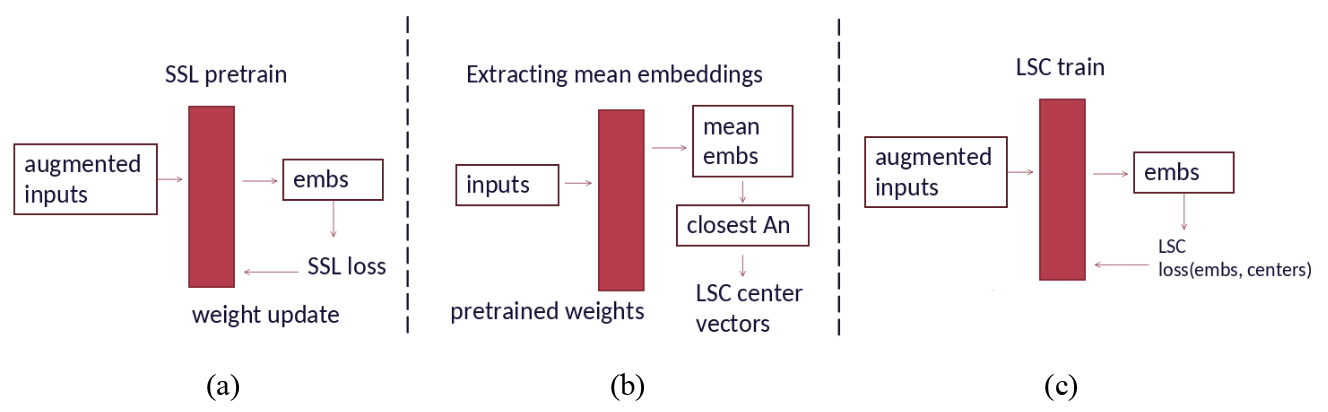} 
	\caption{The proposed multistage SSL-LSC training to obtain target
	configuration close to the preferred distribution: (a) pretraining model
	from scratch using SSL, (b) calculating training dataset mean embeddings
	using model weights obtained in (a) to find closest
	\emph{A\textsubscript{n}} vectors, and (c) using center vectors from (b)
	as target configuration for LSC training.}
	\label{fig:661}
\end{figure}

However, obtaining discriminative features which are preferred by NNs on
specific datasets is precisely the aim of SSL methods. Therefore, it can
be proposed to initially train NNs using strong SSL methods to obtain
representative embeddings as shown in Figure~\ref{fig:661} (a). Then center vectors
from the desired distribution (e.g., \emph{A\textsubscript{n}}) can be
chosen as the closest to the mean SSL embeddings (see Fig.~\ref{fig:661} (b)).
This will make it easier for NN to train on this specific distribution.
Moreover, at this point the center-label correspondence can also be
determined assuming dataset labels are available. Finally, the NN is
trained further using LSC methodology proposed in this paper using SSL
center vectors as target embeddings. Therefore, the initial
computational overhead of SLL training is leveraged to significantly
speed up LSC training.

\unskip

\subsection{Remarks regarding LSC training speed}
\label{speed}

Currently, the main drawback of LSC compared to SL methods is slower
training speed. However, it should be kept in mind that conventional
methods that use entropy achieve faster training by assigning specific
neurons to classes. On one hand that makes finding optimal weights
easier, while on the other hand it associates NN parameters with classes
leading to larger NNs required to accommodate more classes. LSC forfeits
this option in favor of achieving other advantages discussed in Sections~\ref{nodep}
and~\ref{lifelong}. Therefore, while speed-wise LSC cannot compete with CE and
other similar methods in conventional cases, LSC is promising for cases
where conventional methods cannot be applied.

However, it should be kept in mind that the LSC methodology is still at
a relatively early stage of its development. Hence, there is a high
chance that effective training approaches will be found in the future.
The history of NN research has seen many examples of combined loss
functions effectively addressing the drawbacks of losses that constitute
them. Future work will focus on researching new techniques that allow
faster convergence for cosine and embedding matching training methods to
improve existing and facilitate new LSC application in additional
scientific areas.

\section{Conclusions}
\label{conclusions}

This paper formalizes latent space configuration methodology for NN
training which can be used on data with arbitrarily large numbers of
classes. This is achieved by matching NN embeddings with a predefined
embedding distribution with desired properties. Possible NN embedding
distributions are discussed from theoretical and practical standpoints,
and \emph{A\textsubscript{n}} root system vectors are chosen as the
target distribution for experiments in this paper. LSC applicability is
verified in low- and high-dimensional cases by training encoders and ViT
models on cinic and ImageNet-1K. The absence of the dependence of NN
parameter number of the number of classes during LSC training is then
utilized to train ViT-S and encoders on data with up to 1.28 million classes. 
The experiments verify that the GPU memory required for training is independent 
of the number of classes, which allows to use LSC in cases when using 
conventional method becomes unfeasible or even impossible. 
It is discussed that the main disadvantage of LSC is slower training speed
compared to other SL methods, and potential research directions such as
SSL-LSC combined methodology to determine the preferred NN distribution
for faster LSC training are outlined. Additional discussions include
potential applications of LSC in NN distillation and lifelong learning
illustrating the versatility of the proposed methodology.

\section*{Acknowledgement}
\label{acknowledgement}

The author would like to thank his colleagues Dr Anton Raskovalov, Dr
Igor V. Netay, and Ilya Androsov for fruitful discussions, and Vasily
Dolmatov for discussions and project supervision.


\bibliographystyle{IEEEtran}
\bibliography{IEEEabrv,ms}

\end{document}